\documentclass[times,twocolumn,final]{elsarticle}

\usepackage{geometry}
\pdfoutput=1

\usepackage{framed,multirow}

\usepackage{amssymb}
\usepackage{latexsym}

\usepackage{url}
\usepackage{xcolor}
\usepackage{hyperref}

\definecolor{newcolor}{rgb}{.8,.349,.1}

\usepackage{hyperref}
\usepackage{tcolorbox}
\usepackage{framed}
\usepackage{pifont}
\usepackage{amssymb}
\usepackage{amsmath}

\usepackage{enumitem}
\newcommand{\subscript}[2]{$#1 _ #2$}

\usepackage{graphicx}
\usepackage{url}
\usepackage{listings}
\usepackage{bm}
\usepackage{array}
\usepackage{color}
\usepackage{threeparttable}
\usepackage{float}
\usepackage{wrapfig}
\usepackage{xcolor}
\usepackage{tikz}
\usepackage{makecell}
\usepackage{cuted}
\usepackage{capt-of}
 \definecolor{darkblue}{rgb}{0,0.08,0.45}
\definecolor{shadecolor}{rgb}{0.95,0.95,0.92}
\usepackage{xcolor}
\usepackage{xcolor,colortbl}
\usepackage{listings,ulem}
\newtheorem{assumption}{\bf Assumption}
\usepackage{algorithm}
\usepackage{algpseudocode}
\definecolor{lightblue}{RGB}{240,245,255}
\definecolor{codegray}{rgb}{0.5,0.5,0.5}
\usepackage[framemethod=tikz]{mdframed}
\definecolor{shadecolor}{RGB}{240,245,255}
\usepackage{amsmath}
\usepackage{tikz}

\usetikzlibrary{shapes,arrows,arrows,positioning,matrix}
\urldef{\mailsc}\path|matliuj@nus.edu.sg|

\usepackage{tikz}

\begin{document}


\begin{frontmatter}

\title{Compressed Sensing Plus Motion (CS+M): A New Perspective for Improving Undersampled MR Image Reconstruction }

\author[1]{Angelica I Aviles-Rivero \corref{cor1}}
\cortext[cor1]{Corresponding author:
  ai323@cam.ac.uk;}
\author[2]{No\'emie  Debroux}
\author[3]{Guy  Williams}
\author[4]{Martin J.   Graves}
\author[5]{Carola-Bibiane Sch\"{o}nlieb}

\address[1]{Department of Pure Mathematics and Mathematical Statistics, University of Cambridge, UK}
\address[2]{Universit\'e Clermont Auvergne, CNRS, SIGMA Clermont, Institut Pascal, France}
\address[3]{Wolfson Brain Imaging Centre, Department of Clinical Neurosciences, University of Cambridge, UK}
\address[4]{Department of Radiology, Cambridge University Hospitals,  University of Cambridge, UK}
\address[5]{Department of Applied Mathematics and Theoretical Physics, University of Cambridge, UK}

\begin{abstract}
We address the problem of reconstructing high quality images from undersampled MRI data. This is a challenging task due to the highly ill-posed nature of the problem. In particular, in dynamic MRI scans, the interaction between the target structure and the physical motion affects the acquired measurements \textcolor{black}{leading to blurring artefacts and loss of fine details}.
In this work, we propose a framework for dynamic MRI reconstruction framed under a new multi-task optimisation model called Compressed Sensing Plus Motion (CS+M). Firstly, we propose a single optimisation problem that simultaneously computes the MRI reconstruction and the physical motion. Secondly, we show our model can be efficiently solved by breaking it up into two more computationally tractable problems. The potentials and generalisation capabilities of our approach are demonstrated in different clinical applications including cardiac cine, cardiac perfusion and brain perfusion imaging. We show, through numerical and graphical experiments, that the proposed scheme reduces blurring artefacts and preserves the target shape and fine details. We also report the highest quality reconstruction under highly undersampling rates in comparison to several \textcolor{black}{state of the art} techniques.
\end{abstract}


\end{frontmatter}


\section{Introduction}
A central limitation of  Magnetic Resonance Imaging (MRI) is the linear relation between the number of necessary measurements to form an image and the acquisition time. This constraint causes negative effects including~\citep{havsteen2017movement,Hollingsworth2015,Zaitsev::2015}: (i) sensitivity to motion \textcolor{black}{responsible for} image degradation, (ii) reduced clinical throughput  and (iii) patient non-compliance, \textcolor{black}{ generating} 
more artefacts during the image formation. 
Thus, a central challenge in MRI is to respond to the question $-$ How to decrease the long data-acquisition time?

There have been different attempts to answer this question, 
which \textcolor{black}{{solutions}} can be divided into two categories: hardware-based or software-based approaches. 
\textcolor{black}{The former ones }aim to redesign the internal mechanisms of the scanner. However, despite the continuous development of technologies, at present, there is no evidence that the MRI data-acquisition time problem can be solved using a hardware-based approach~\citep{Majumdar::2015,Zaitsev::2015}. This is primarily due to both the physical limitations of gradient hardware, and the physiological limits imposed for safe pulse sequences.

Unlike hardware-based approaches, the second category has shown significant potential for effectively reducing acquisition times. 
This is particularly due to the advent of Compressed Sensing (CS) in MRI~\citep{Lusting::2007}, which is motivated by the solid mathematical foundations of CS~\citep{CandesStable2006,Donoho::2006}. The key idea of CS in MRI is to form an image - represented in an appropriate transform basis - from a considerably reduced finite-dimensional subset of \textit{k}--space data acquired in an incoherent manner \textcolor{black}{ and relying on sparsity principles}.

Notwithstanding that the body of literature has evidenced powerful results when using CS 
methods for MR image reconstruction, in what follows we show that there still is significant room for improvement in terms of reconstruction quality in the context of spatio-temporal MRI reconstruction. Namely, in this work we introduce a mathematical framework  to improve undersampled MR image reconstruction in a dynamic setting which we call \textit{Compressed Sensing plus Motion} (CS+M). 

The central idea of our CS+M model is to compute, \textit{in a single model}, a compressed sensing for fast MRI reconstruction loss, and an energy model derived from the complex motion patterns in the scene (i.e. from physiological or involuntary motion). More precisely, we frame the image reconstruction task as a unified minimisation problem, which is efficiently solved by an alternating scheme.
Whilst this is an important part of our solution, our main contributions are:

\begin{itemize}
  \item We propose a computationally tractable variational model framed as a multi-task framework, in which we highlight:
    \begin{itemize}
        \item A single optimisation model that allows establishing, explicitly and simultaneously, the connection between  MR image reconstruction and the \textcolor{black}{estimation of the} inherent physical motion captured during the data acquisition.
        \item \textcolor{black}{An efficient alternating minimisation technique to solve the original optimisation problem.}
    \end{itemize}
  \item  We show that incorporating physical motion into the compressed sensing computation improves the MR image reconstruction from undersampled data, and produces images closer to fully sampled data whilst requiring less measurements. Furthermore, we show that image quality persists for higher acceleration factors than can be tolerated if motion is not included in the model.
 \item We validate the feasibility and generalisation capabilities of our model, through numerical and graphical results, with several clinical applications including cardiac \textcolor{black}{cine}, cardiac perfusion and brain perfusion data.
\end{itemize}


\section{Related work}
The problem of MRI reconstruction from highly undersampled dynamic measurements has been widely investigated in the community. In this section, we review the existing techniques in this context.

The potential and benefits of using CS in MRI have been demonstrated in different works starting with the pioneering work by Lustig et al. in~\citep{Lusting::2007}, and 
\textcolor{black}{followed} by different approaches such as k-t SPARSE~\citep{LustigktSPARSE::2006}, k-t SPARSE-SENSE~\cite{Otazo::2010} and L1-SPIRiT~\citep{LustigSPIRiT::2010}, in which CS implications have been applied alone or in combination with parallel imaging.

A different direction based on low-rank matrix completion, which extends the idea of CS, has gained the attention of other scientists \textcolor{black}{who} 
have explored the effects of applying either local or global low-rank constraints.
Liang in~\citep{Liang::2007}  proposed to compute temporal basis functions, using singular value decomposition, as a new form to achieve MR image reconstruction from undersampled \textit{k\textcolor{black}{,t}-}space data. This work set\textcolor{black}{s} the basis and the motivation in the use of principal component analysis (PCA) for reconstructing a small subset of \textit{k\textcolor{black}{,t}}-space data (e.g.~\citep{Feng::2013,Pedersen::2009,Velikina::2015}). The idea of computing locally low-rank (LLR) constraint, in the context of undersampled MR image reconstruction, was reported in~\citep{Miao::2016,Trzasko::2011,Zhang::2015}, in which spatiotemporal correlations were analysed in small regions. Although LLR has been suggested to decrease computational load~\citep{Trzasko::2011}, in comparison to global low-rank constraint, it comes at the expense of dealing with block artefacts~\citep{Saucedo::2017}.

The concept of combining sparse and low-rank constraints was also successfully reported by the MRI community. The authors in~\citep{Lingala::2011,Majumdar::2012,Zhao::2012} aimed to find a solution that is both sparse and low-rank. \textcolor{black}{Also, the BCS approach was introduced in \citep{lingala::2013} and aimed at simultaneously estimat\textcolor{black}{ing} the temporal basis functions with no orthogonal constraints as well as the sparse coefficients from the undersampled measurements.} A different approach based on decomposing the acquired measurements as a linear combination of low-rank (L) and sparse (S) components $-$ known as Robust Principal Component Analysis (RPCA)~\citep{Candes::2011} or L+S decomposition $-$  was investigated in~\citep{Gao::2012,Tremoulheac::2014,Otazo::2013,Otazo::2015,Singh::2015}. Particularly Otazo et al. demonstrated the feasibility of L+S model in~\citep{Otazo::2015} for various clinical applications in MRI including cardiac \textcolor{black}{cine} and abdominal perfusion. Even though the L+S model has proved to be effective for MR image reconstruction, the separation of background and dynamic components, in fact, is not always possible since the incoherence required by this model, i.e. for L and S, is not always fulfilled.

Another line of search in dynamic MRI relies on the observation of motions patterns in such acquisitions e.g.~\citep{Bilen::2011,Usman::2013}.
For example authors of~\citep{Royuela::2017} introduced a new regularisation metric based on a spatial weighting given by the Jacobian of the estimated deformation via a groupwise B-spline parametric deformable registration technique, to improve the reconstruction in a compressed sensing framework. Rank et al. \citep{Rank::2017} proposed an alternating scheme, in a coarse-to-fine setting, between motion-compensated image reconstruction based on HDTV and artefact robust motion estimating relying on the Demons algorithm and a cyclic self-consistency constraint, for 4D respiratory time-resolved MRI.  


Whilst 
\textcolor{black}{most of the literature} seek\textcolor{black}{s} to solve only the reconstruction task or \textcolor{black}{to} include motion \textcolor{black}{information} in a sequential fashion, our work is more closely related to multi-task (join\textcolor{black}{t} models) approaches. The central idea of this perspective is to tackle, simultaneously and explicitly, two or more task\textcolor{black}{s} that share a common structure~\citep{caruana1997multitask}. In this work, our task\textcolor{black}{s} are reconstruction and motion estimation. These task\textcolor{black}{s} ha\textcolor{black}{ve} been recognised as highly related to each other since early developments in MRI \citep{Wood::85,Van::97}. Our \textcolor{black}{main motivation for designing} a multi-task approach is to reduce the error propagation and to take advantage of the beneficial mutual influence of these two tasks.

Following the multi-task perspective different works have been reported. Authors of
\citep{Odille::2016} proposed a  joint optimisation framework correcting for both inter- and intra-image motion requiring motion sensors. The Deformation Corrected - Compressed Sensing (DC-CS) model was introduced in \citep{Lingala::2015}. This model relied on compressed sensing techniques along with a Demons registration method, and was solved in an alternating way after introducing auxiliary variables. The philosophy of our work is similar to the works of that \citep{Burger::2018,Zhao::2019}.  Motivated by the combined principles of compressed sensing and optical flow, \textcolor{black}{we} emphasize that the main differences stand in the way of treating the optical flow task, which \textcolor{black}{is} described next.

\textcolor{black}{In \citep{Burger::2018}, authors provide} a theoretical analysis of a general formulation, while in this work, we establish a solid proof of concept that jointly addressing highly undersampled MRI reconstruction and motion estimation, in the original and complete optical flow setting, improves the quality of the recovered images. \textcolor{black}{We take this model as backbone and shape it according to our specific application.}
Whilst the authors of~\cite{Zhao::2019} simplified the weighted optical flow task by considering only affine displacement fields, 
in our model, 
\textcolor{black}{it} is not restricted to any particular class of functions and we consider the un-weighted optical flow formulation.

\section{MRI Reconstruction: Preliminaries}

In a dynamic MRI setting, the signal measurements, $\mathbf{y}$, are collected in a time-spatial-frequency space (i.e. \textit{k,t}-space) and expressed as $y(\mathbf{k},t)=\int_\Omega u(\mathbf{x},t)\mbox{exp}(-j\mathbf{k}^T\mathbf{x})\,dx + \mathbf{\eta}(\mathbf{k},t)$, where $\Omega$ is a connected open bounded subset of $\mathbb{R}^2$ representing the definition domain of the image and $\mathbf{x}$ the 2D spatial coordinate. Moreover, $\mathbf{k}$ is the 2D frequency variable, $t$ denotes the temporal coordinate in $[0,T]$ with $T$ \textcolor{black}{being} the end time of acquisition, $\mathbf{\eta}$ is inherent noise during the acquisition, and $u$ is the target stack of images showing a moving part of the patient body over the acquisition time.



A current research direction in MRI is focused on reconstructing a small number of measurements with the aim of decreasing the acquisition time. That is, $y(\mathbf{k},t)$ is available for only a small amount of values of $\mathbf{k}\in \mathbb{R}^2$ for each time $t$. However, the inverse reconstruction problem becomes highly underconstrained and can only be achieved by adding a prior information on the reconstruction. The implicit sparsity of MR images in a transform domain is often used in this prospect~\citep{Lusting::2007}. Based on Compressed Sensing (CS) principles, the $L^1$-norm is used as a measure of sparsity, and the problem of reconstructing undersampled MRI data becomes:

\begin{align}
\underset{u}{\mbox{argmin}} \|\mathcal{T}u\|_{1},\, \text{s.t. } \|\mathcal{A}u-y\|_{2}^2\leq \alpha,
\label{constrained_reconstruction}
\end{align}

where $\mathcal{A}$ denotes  the associated undersampled Fourier operator \textcolor{black}{, also encoding coil sensitivities,} which is a linear, continuous and bounded operator according to Plancherel's theorem. Moreover, $\mathcal{T}$ is the transform domain, in which a sparse representation is promoted $-$ common transforms include Wavelet Tranform~\citep{Lusting::2007} widely used in compression, spatial finite differences such that $\|\mathcal{T}u\|_{1} = TV(u)$ (Total Variation \citep{Rudin::1992}) or $TGV(u)$ (Total Generalised Variation, \citep{Schloegl::2017}). In what follows, we choose the Total Variation operator favoring piecewise constant reconstruction with sharp edges, and $\|\mathcal{T}u\|_1$ is replaced by $TV(u)$. For tractability purposes, the constrained problem ~\eqref{constrained_reconstruction} can be relaxed to the following unconstrained problem using an $L^p$-penalty method, which reads:

\begin{align}
 \label{unconstrained_reconstruction}
 \underset{u}{\mbox{argmin}} \left\{ \frac{1}{2}\|\mathcal{A}u-y\|_2^2 + \gamma \|\mathcal{T}u\|_1\right\},
\end{align}

\noindent
where $\gamma>0$ is a weighting parameter between the fidelity term and the regularisation. Although the body of literature for reconstructing undersampled MRI data has provided promising results \--- with generalised CS approaches such as the scheme described in (\ref{unconstrained_reconstruction}), and extended ideas \cite{Lingala::2011,Majumdar::2012,Otazo::2015}\--- MRI reconstruction is still a challenging and open problem and there is plenty of room for further improvements especially in the dynamic setting. In particular, one seeks to cope with -  \textit{How to reconstruct high-quality MR images with less measurements?} We address this question by proposing a new multi-task framework called CS+M which, unlike \textcolor{black}{most of} existing approaches, considers, explicitly and simultaneously, the computation of the inherent complex motion patterns derived from physiological or involuntary motion. 

\section{Compressed Sensing Plus Motion (CS+M)}
In this section, we introduce our CS+M model for improving under-sampled MRI reconstruction in dynamic settings.


\begin{figure*}[!t]
\centering
\includegraphics[width=1\textwidth]{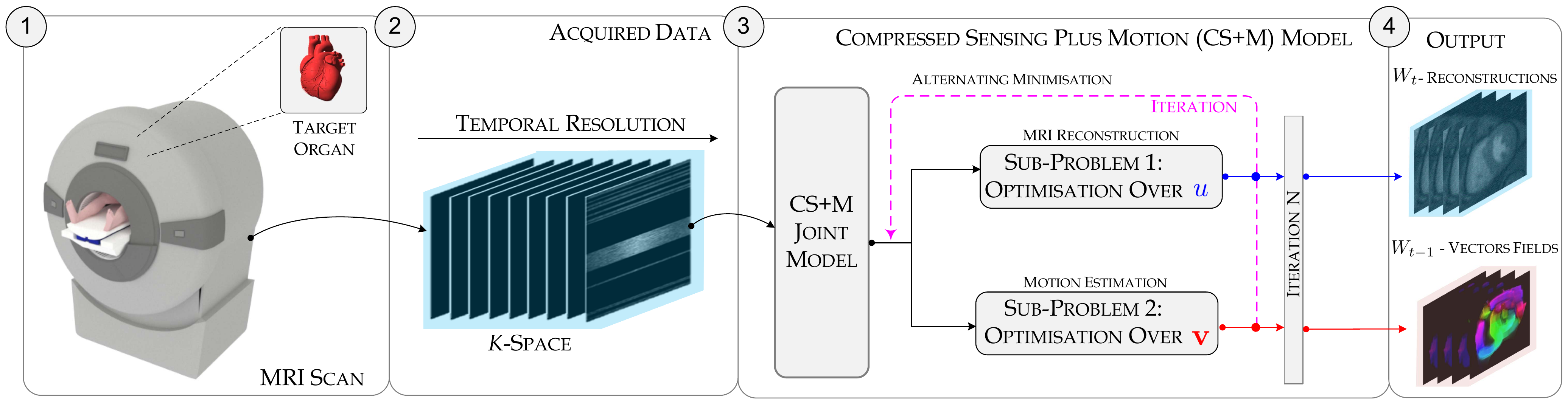}
\caption{From left to right: MRI setting in which spatial-frequency information is acquired to reconstruct a target body's part of the patient. \textcolor{black}{Illustration of our proposed scheme, in which explicit and simultaneous computation of the physical motion is injected in the algorithmic MRI reconstruction}.}
\label{fig::datasets2A}
\end{figure*}

Our algorithmic approach is strongly motivated by the fact that MRI is well-known to be highly sensitive to motion since its early development \cite{Wood::85,Van::97}. In dynamic MRI scans, in particular, the inherent physical motion of the target structure affects the acquired measurements resulting in quality degradation of classical sequential CS reconstructed images which compromises clinical interpretations. Our main motivation is thus:

\begin{assumption}
 Incorporating physical motion estimation into a given dynamic MRI reconstruction method result\textcolor{black}{ing} in a faster acquisition allowing for extreme under-sampling, and in higher time resolution leading to better image quality.
\end{assumption}

The first question we thus need to answer is {\textit{How to model the inherent physical motion in the scene?}}

The problem of estimating motion from a set of images acquired dynamically, has been extensively investigated in the community with \textcolor{black}{models} categorised in direct and indirect methods \citep{Zisserman::1999,Irani::1999}. \textcolor{black}{However} it is still considered as a challenging problem and remains an active research topic. In this work, we focus on the Optical Flow setting, which is one of the most well-established methods starting with the pioneering work of Horn and Schunck \citep{Horn::81}. Optical Flow relies on the linearised {\textit{brightness constancy assumption}} also known as the {\textit{optical flow constraint}}. It links intensity variations in image sequences $u(x,t)$ to the underlying velocity fields $\mathbf{v}(x,t)$ by assuming the image intensity is constant along the trajectory $x(t)$ with $\mathbf{v}(x,t)=\frac{dx(t)}{dt}$ :
$ 0=\frac{du}{dt} = \frac{\partial u}{\partial t} + \frac{\partial u}{\partial x}\frac{\partial x}{\partial t} = u_t+\nabla u.\mathbf{v}$,
where $\nabla u$ is the \textcolor{black}{spatial} gradient image. This problem is underdetermined since it gives only one equation to estimate a bidimensional variable $\mathbf{v}$ for 2D images. However, one can still estimate the motion by introducing prior information on the nature of the velocity field in a variational framework \citep{Horn::81}.

From a variational setting perspective, Optical Flow can be cast as the problem of minimising
an energy functional with respect to the velocity field $\mathbf{v}$, composed of a fidelity term $\mathcal{D}_{\text{Motion}}$ measuring the brightness constancy deviations, and a regularisation term $\mathcal{J}_{\text{Motion}}$ penalising high variations in $\mathbf{v}$. Formally, one seeks to solve the following functional:

\begin{align}
\label{eq:OF}
 \nonumber E_{OF}(u, \mathbf{v})&=\int_0^T\mathcal{D}_{\text{Motion}}(u,\mathbf{v})+\delta \mathcal{J}_{\text{Motion}}(\mathbf{v})\,dt,\\
 &=\int_0^T\|\nabla u. \mathbf{v} + u_t \|_1+\delta TV(\mathbf{v})\,dt,
\end{align}

\noindent
where $\delta>0$ is a weighting parameter balancing the influence of each term. Our motivation for choosing the $TV-L^1$ optical flow model is twofold. Firstly, we aim to gain robustness in terms of outliers, which is important for MRI  as inherent noise during MRI acquisition deteriorates the quality of the reconstruction. Secondly, this formulation allows for piecewise constant flow fields with discontinuities. This model has been studied in \citep{Garg::2011,Zach::2007,Perez::2013} and it proves to give a robust and good estimate of the physical motion.

We now turn to detail how \eqref{unconstrained_reconstruction} and \eqref{eq:OF} can be carefully intertwined. In particular, we describe our multi-task variational model for simultaneously reconstructing dynamic MR images as well as estimating the underlying physical motion, exploiting then both spatial and corrected temporal redundancy to further improve image quality. 
\textcolor{black}{T}he CS+M model can be cast as an optimisation problem, in which one seeks to optimise over the MRI reconstruction $u$ and the estimated motion $\mathbf{v}$:


\begin{align}
 \nonumber&\underset{u,\mathbf{v}}{\mathrm{argmin}} \int_0^T \bigg\{ \underbrace{\mathcal{D}_{\text{rMRI}}(u(.,t))}_{\begin{array}{l}\text{\small{Data fidelity}}\\\text{\small{Reconstruction}}\end{array}}+\gamma \underbrace{\mathcal{R}_{\text{rMRI}}(\mathcal{T}u(.,t))}_{\begin{array}{l}\text{\small{Regularisation promoting}}\\\text{\small{spatial sparsity}}\end{array}}\\
\nonumber &+\beta\underbrace{\mathcal{D}_{\text{Motion}}(u(.,t),\mathbf{v}(.,t))}_{\begin{array}{l} \text{\small{Data fidelity}}\\\text{\small{Physical motion}}\end{array}}+\delta \underbrace{\mathcal{J}_{\text{Motion}}(\mathbf{v}(.,t))}_{\begin{array}{l}\text{\small{Regularisation}}\\\text{\small{Physical motion}}\end{array}}\,dt\bigg\},\\
\nonumber  \Leftrightarrow&\, \underset{u,\mathbf{v}}{\mathrm{argmin}}\int_0^T \bigg\{ \frac{1}{2}\|\mathcal{A}u(.,t)-y(.,t)\|_2^2+\gamma TV(u(.,t))\\
&+\beta \|\nabla u (.,t). \mathbf{v}(.,t) + u_t(.,t)\|_1 +\delta TV(\mathbf{v}(.,t))\bigg\}\,dt,  \label{initial_problem}
\end{align}

\noindent
with $\beta>0,\, \gamma>0,\, \delta>0$, positive weighting parameters balancing the influence of each term.

The purpose of this work is a proof-of-concept study to show that spatio-temporal approach estimating explicitly and simultaneously the physical motion can generally improve the image quality upon a simple frame-by-frame reconstruction in an MRI setting. We show in the next section, that this improvement happens \textcolor{black}{also} for perfusion MRI with very promising results even if the brightness constancy is not fulfilled in this context.
We follow a discretise then optimise strategy to numerically solve problem (\ref{initial_problem}).


\subsection{Optimisation Scheme}
For tractability purposes, we define a spatio-temporal-discrete version of (\ref{initial_problem}) by performing the discretisation of the time interval [0,T] into $W_t$-steps and of the spatial domain $\Omega$ into the pixel grid $\{1,\cdots,M\}\times\{1,\cdots,N\}$. This result\textcolor{black}{s} in $W_t$ reconstructed images $u$ and in $W_t-1$ vector fields $\mathbf{v}$ on the pixel grid. We then seek to minimise the following functional:

\noindent
\begin{align}
  \nonumber &\underset{u,\mathbf{v}}{\mathrm{argmin}}\Bigg\{ \underset{k=1}{\overset{W_t}{\sum}}\frac{1}{2}\|\mathcal{A}u^k-y^k\|_2^2+\gamma \|\nabla u^k\|_{2,1} \\
  &+\underset{k=1}{\overset{W_t-1}{\sum}} \delta (\|v_1^k\|_{2,1}+\|v_2^k\|_{2,1})+\beta \|\nabla u^k.\mathbf{v}^k+u_t^k\|_1\Bigg\}, 
 \label{discretised_problem}
\end{align}

\noindent
where $TV$ has been replaced by the discrete isotropic $TV$: $\|\nabla u\|_{2,1} := \underset{i=1}{\overset{M}{\sum}}\underset{j=1}{\overset{N}{\sum}}|\sqrt{u_x(i,j)^2+u_y(i,j)^2}|$.
Due to the scalar product $\nabla u^k. \mathbf{v}$, the energy is nonconvex with non-differentiable $L^1$ norms involved making the numerical resolution challenging. Therefore, to achieve more computational tractability, we propose an alternating minimisation scheme to solve (\ref{discretised_problem}). We thus break up the problem into two more tractable sub-problems, in which we fix one of the variable\textcolor{black}{s} and minimise the functional with respect to the other one alternatively.
This leads to the following iterative two-step scheme in which $u_l=(u_l^k(i,j))_{i=1,\cdots,N,j=1,\cdots,M,k=1,\cdots,W_t}\in \mathbb{R}^{NMW_t}$ and $v_{1,l}=(v_{1,l}^k(i,j))_{i=1,\cdots,N,j=1,\cdots,M,k=1,\cdots,W_t-1}\in \mathbb{R}^{NM(W_t-1)}$, $v_{2,l}=(v_{2,l}^k(i,j)_{i=1\cdots,N,j=1,\cdots,M,k=1,\cdots,W_t-1}\in \mathbb{R}^{NM(W_t-1)}$ :

\smallskip
\textcolor{darkblue}{\ding{42} {\textbf{Sub-problem 1: Optimisation over $u$}}}. The optimisation problem over $u$ while keeping $v_{1,l}$, and $v_{2,l}$ fixed is defined as
 \begin{align}
\nonumber& u_{l+1}=\underset{u}{\mathrm{argmin}}\bigg\{ F(u)=\underset{k=1}{\overset{W_t}{\sum}} \frac{1}{2}\|\mathcal{A}u^k-y^k\|_2^2\\
&+\gamma \|\nabla u^k\|_{2,1}+ \beta \underset{k=1}{\overset{W_t-1}{\sum}} \|u_x^kv_{1,l}^k+v_{2,l}^ku_y^k+u_t^k\|_1\bigg\},
 \end{align}

\noindent
where
$$u_x^k(i,j)=D_xu^k(i,j)=\left\{
                        \begin{array}{cc}
                        \frac{u^k(i+1,j)-u^k(i-1,j)}{2} &\text{ if } 1<i<N,\, k<W_t,\\
                          0&\text{ else,}
                         \end{array}
                         \right.$$
$$u_y^k(i,j)=D_yu^k(i,j)=\left\{\begin{array}{ll} \frac{u^k(i,j+1)-u^k(i,j-1)}{2} &\text{ if } 1<j<M,\, k<W_t,\\ &\text{ else,}\end{array} \right.$$
\noindent
and
$$u_t^k(i,j)=D_tu^k(i,j)=\left\{ \begin{array}{ll} u^{k+1}(i,j)-u^k(i,j) &\text{ if }k<W_t,\\0 &\text{ else}, \end{array}\right.$$

\noindent
This is a typical ROF problem \citep{Rudin::1992} coupled with a transport term coming from the optical flow component. We propose to use a primal-dual algorithm to solve this problem \citep{Chambolle::2011}. We thus introduce the following discrete operator:
\begin{align*}
 C_uu = \begin{pmatrix} \mathcal{A} \\ \nabla \\ D_t+v_{1,l}D_x+v_{2,l}D_y  \end{pmatrix}u,
\end{align*}
and compute the convex conjugate $F^*$ corresponding to $F$ as:
\begin{align}
 \nonumber F^*(\mathbf{y})&=\underset{k=1}{\overset{W_t}{\sum}}\frac{1}{2}\|y_1^k\|_2^2+\langle y_1^k,y^k\rangle+\mathrm{1}_{\{y:\|y\|_{2,\infty}\leq1\}}(\frac{y_2^k}{\gamma})\\
 &+\mathrm{1}_{\{y:\|y\|_\infty\leq 1\}}(\frac{y_3^k}{\beta}).
\end{align}
This leads to the following primal-dual problem, which reads:
\begin{align}
 \nonumber &\underset{u}{\mathrm{argmin}}\,\underset{\mathbf{y}}{\mathrm{argmax}} \underset{k=1}{\overset{W_t}{\sum}} \langle C_uu^k,\mathbf{y}^k\rangle -\frac{1}{2}\|y_1\|_2^2-\langle y_1^k,y^k\rangle \\
 &- \mathrm{1}_{\{y:\|y\|_{2,\infty}\leq 1\}}(\frac{y_2^k}{\gamma})-\mathrm{1}_{\{y:\|y\|_\infty\leq 1\}}(\frac{y_3^k}{\beta}),
\end{align}
and the following iterative primal-dual scheme :
\begin{align}
 \tilde{\mathbf{y}}_{h+1}&=\mathbf{y}_h+\sigma C_u(2u_h-u_{h-1}),\\
 y_{1,h+1}&=\frac{\tilde y_{1,h+1}-\sigma y}{\sigma+1},\\
 y_{2,h+1}&=\pi_\gamma(\tilde y_{2,h+1}),\\
 y_{3,h+1}&=\pi_\beta(\tilde y_{3,h+1}),\\
 u_{h+1}&=u_l-\tau C_u^T\mathbf{y}_{h+1},
\end{align}
with $\pi_\alpha(y)=\frac{y}{\max(1,\frac{\|y\|_2}{\alpha})}$, $C_u^T$ the adjoint operator, and $\sigma$ and $\tau$ step sizes. After convergence of this iterative scheme, we set $u_{l+1}=u_{h}$.

\medskip
\textcolor{darkblue}{\ding{42} {\bf{Sub-problem 2 : Optimisation over $\mathbf{v}$}}}. The optimisation problem over $\mathbf{v}$ while keeping $u_l$ fixed is defined as:
\begin{align}
 \nonumber &\underset{\mathbf{v}=(v_1,v_2)}{\mathrm{argmin}} \underset{k=1}{\overset{W_t-1}{\sum}}\|\nabla u_{l+1}^k.\mathbf{v}^k+u_{t,l+1}^k\|_1+\frac{\delta}{\beta}\underset{k=1}{\overset{W_t}{\sum}}(\|\nabla v_1\|_{2,1}\\
 &+\|\nabla v_2\|_{2,1}).
\end{align}
This is a classical $L^1-TV$ optical flow problem which can also be solved by a primal-dual algorithm. We thus formulate the following associated primal-dual problem as:
\begin{align}
 \nonumber&\underset{\mathbf{v}}{\mathrm{argmin}}\,\underset{\mathbf{z}}{\mathrm{argmax}} \underset{k=1}{\overset{W_t}{\sum}} \|u_{t,l+1}+\nabla u_{l+1}^k.\mathbf{v}\|_1+\langle C_v\mathbf{v},\mathbf{z}\rangle\\
 &-\mathrm{1}_{\{y:\|y\|_{2,\infty}\leq 1\}}(\frac{z_1\beta}{\delta})-\mathrm{1}_{\{y:\|y\|_{2,\infty}\leq 1\}}(\frac{v_2\beta}{\delta}),
\end{align}

\begin{algorithm}[t!]
\label{alg::ours}
\caption{Our Proposed Method}
\begin{tcolorbox}[colback=lightblue, colframe=white]
\vspace{-0.3cm}
\begin{lstlisting}[language=Python, mathescape=true]
Start from $u \leftarrow 0$ and $\mathbf{v} \leftarrow 0$;
Set $\zeta_{stop} \leftarrow 10^{-5}$;
while $d_{error}>\zeta_{stop}$:
    $u_{previous} \leftarrow u$;
    $\mathbf{v}_{previous} \leftarrow \mathbf{v}$;
    $y,\bar{u},\bar{\mathbf{v}},\mathbf{z}\leftarrow 0$;
    while {$ \epsilon_u > \zeta_{stop}$}:
        $u_{Old} \leftarrow u$;
        $\tilde{\mathbf{y}} \leftarrow \mathbf{y}+\sigma C_u \bar{u}$;
        $y_1\leftarrow \frac{\tilde y_1-\sigma y}{\sigma+1}$;
        $y_2=\pi_\gamma (\tilde y_2) $;
        $y_3=\pi_\beta(\tilde y_3)$;
        $u\leftarrow u-\tau C_u^T \mathbf{y}$;
        $\bar{u}\leftarrow 2u-u_{Old}$;
        $\epsilon_u\leftarrow$ solve(absdiff($u_{Old},u$));
    $du_{error} \leftarrow$ solve(absdiff($u_{previous},u$));
    while{$\epsilon_v >\zeta_{stop}$}:
        $\mathbf{v}_{Old} \leftarrow \mathbf{v}$;
        $\tilde{\mathbf{z}} \leftarrow \mathbf{z}+\sigma C_v\bar{\mathbf{v}}$;
        $\mathbf{z} \leftarrow \pi_{\frac{\delta}{\beta}}(\tilde{\mathbf{z}})$;
        $\tilde{\mathbf{v}} \leftarrow \mathbf{v}-\tau C_v^T\mathbf{z}$;
        $\mathbf{v}\leftarrow \tilde{\mathbf{v}}+\left\{ \begin{array}{l} \tau \mathbf{\beta} \text{ if } \rho(\tilde{\mathbf{v}})<-\tau\|\mathbf{\beta}\|_2^2\\-\tau \mathbf{\beta} \text{ if } \rho(\tilde{\mathbf{v}})>\tau\|\mathbf{\beta}\|_2^2\\ -\frac{\rho(\tilde{\mathbf{v}})}{\|\mathbf{\beta}\|_2^2}\mathbf{\beta}\text{ else,}
        \end{array}
        \right. $;
        $\bar{\mathbf{v}}\leftarrow 2\mathbf{v}-\mathbf{v}_{Old}$;
        $\epsilon_v\leftarrow$ solve(absdiff($\mathbf{v},\mathbf{v}_{Old}$));
    $dv_{error}\leftarrow $ solve(absdiff($\mathbf{v}_{previous},\mathbf{v}$));
    $d_{error}\leftarrow $ solve(sum($du_{error},dv_{error}$));
return $u, \mathbf{v}$.
\end{lstlisting}
\vspace{-0.4cm}
\end{tcolorbox}
\vspace{-0.3cm}
\end{algorithm}

\noindent
where $C_v\mathbf{v}=\begin{pmatrix} \nabla & 0\\0 & \nabla \end{pmatrix}\mathbf{v}$ is a discrete operator such that:

$$v_{1,x}^{k}(i,j)=\left\{ \begin{array}{ll} v_1^k(i+1,j)-v_1^k(i,j) &\text{ if } i<N \\ 0 &\text{ if } i=N, \end{array}\right.$$

$$v_{1,y}^k=\left\{ \begin{array}{ll} v_1^k(i,j+1)-v_1^k(i,j) &\text{ if } j<M \\ 0 &\text{ if } j=M, \end{array} \right.$$

$$v_{2,x}^{k}(i,j)=\left\{ \begin{array}{ll} v_2^k(i+1,j)-v_2^k(i,j) &\text{ if } i<N \\ 0 &\text{ if } i=N,\end{array}\right.$$

$$v_{2,y}^k=\left\{ \begin{array}{ll} v_2^k(i,j+1)-v_2^k(i,j) &\text{ if } j<M,\\ 0 &\text{ if } j=M \end{array} \right.$$

Similarly, applying the primal-dual algorithm leads to the following iterative scheme:
\begin{align}
 \mathbf{z}_{h+1} &= \pi_{\frac{\delta}{\beta}}(\mathbf{z}_h + \sigma C_v(2\mathbf{v}_h-\mathbf{v}_{h-1})),\\
\tilde{\mathbf{v}}_{h+1}&=\mathbf{v}_h-\tau C_v^T \mathbf{z}_{h+1},\\
\mathbf{v}_{h+1}&=\tilde{\mathbf{v}}_{h+1}+\left\{ \begin{array}{l} \tau \mathbf{\beta} \text{ if } \rho(\tilde{\mathbf{v}}_{h+1})<-\tau\|\mathbf{\beta}\|_2^2,\\-\tau \mathbf{\beta} \text{ if } \rho(\tilde{\mathbf{v}}_{h+1})>\tau\|\mathbf{\beta}\|_2^2,\\ -\frac{\rho(\tilde{\mathbf{v}}_{h+1})}{\|\mathbf{\beta}\|_2^2}\mathbf{\beta}\text{ else,} \end{array} \right.
\end{align}
with $C_v^T$ being the adjoint operator, $\rho(\tilde{\mathbf{v}}_{h+1})=u_{t,l+1}+\nabla u_{l+1}.\tilde{\mathbf{v}}_{l+1}$, and $\mathbf{\beta}=(u_{x,l+1},u_{y,l+1})^T$. After convergence of this scheme, we set $\mathbf{v}_{l+1}=\mathbf{v}_h$.
The overall procedure is summarised in \ref{alg::ours}.

\section{Experimental Results}
This section describes in detail\textcolor{black}{s} the experiments that we conducted to validate our CS+M reconstruction technique.

\subsection{Data Description}
We evaluate our approach extensively and prove its generalisation by using data coming from: \textcolor{black}{cine} cardiac, cardiac perfusion and brain perfusion MR imaging. The datasets \textcolor{black}{are} saved as fully sampled raw data and then \textcolor{black}{are} retrospectively undersampled using a variable-density random sampling pattern as suggested by Lustig in~\cite{Lusting::2007} and using cartesian sampling. Datasets characteristics are as follows:

\begin{figure*}[htbp]
\centering
\includegraphics[width=1\textwidth]{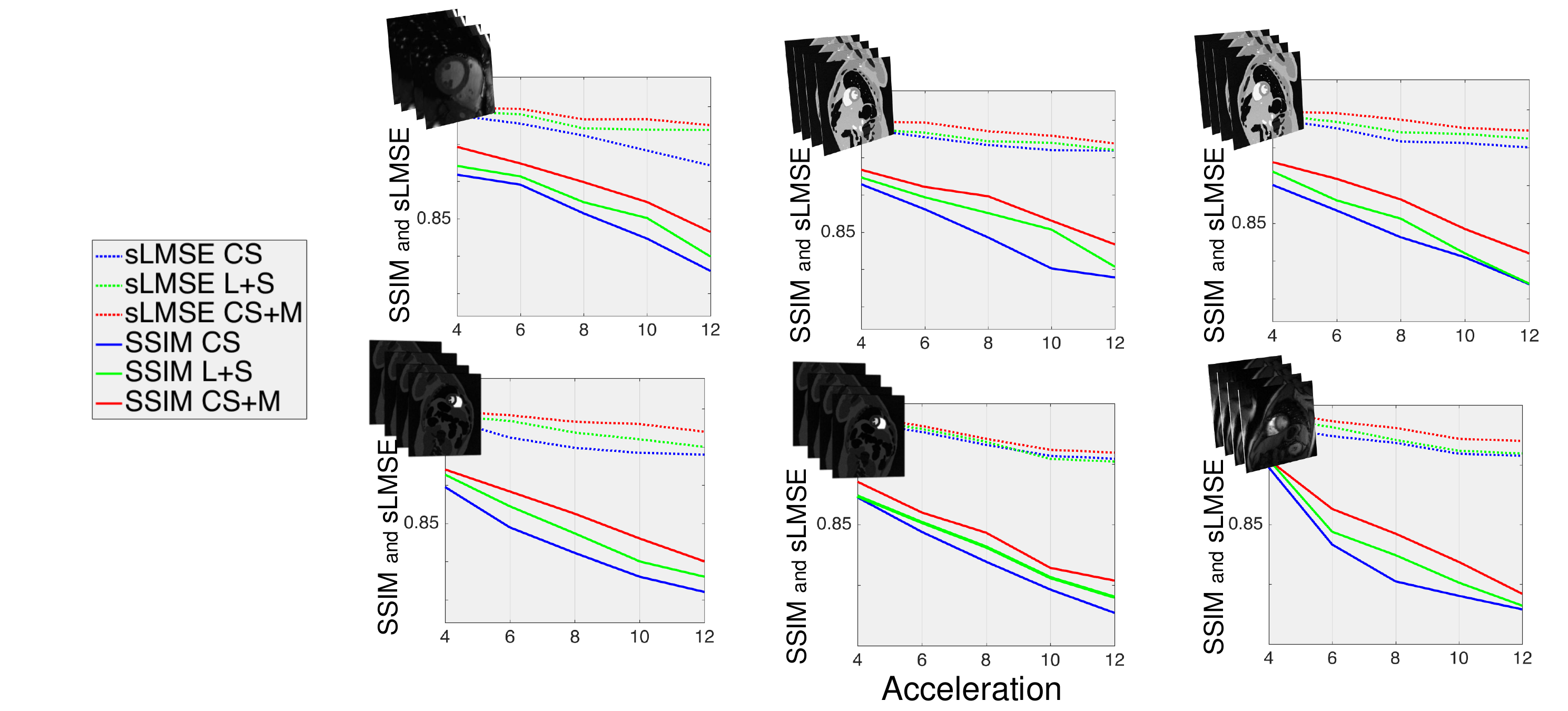}
\caption{Plots comparing CS, \textcolor{black}{L+S} and CS+M using two reconstruction quality metrics, SSIM and sLMSE, over the whole cardiac perfusion datasets.}
\label{fig::Drawing5}
\end{figure*}

\begin{itemize}
\itemsep0em
  \item Dataset I $-$ A \textcolor{black}{cine} cardiac dataset which was acquired from a healthy volunteer, from \citep{Schloegl::2017}. Measurements were collected using a 3T Siemens scanner. Matrix size$- 208\times168$, heart phases$-25$, coils$-30$, FOV = $274.62\times340$mm and TA=16s.
	\item Datasets II and III $-$ Realistic cardiac \textcolor{black}{cine} simulation generated using the MRXCAT phantom framework~\citep{Wissmann::2014}. Whilst the Dataset II was simulated during breath-holding, the III was set with free respiratory motion. The duration of the heart beat and respiratory cycles were set as 1 sec and 5 sec respectively. Both were generated with Matrix size$- 409\times409$,  heart phases$-24$, coils$-12$.
	\item Datasets IV and V $-$ Realistic cardiac perfusion simulations using MRXCAT~\citep{Wissmann::2014} framework. Simulations were set with Matrix size$- 224\times192$,  Frames $-32$, coils$-12$. The difference, for analysis purposes, relies on the fact that the Dataset IV was simulated with breath-holding whilst the dataset V during free-breathing with the respiratory cycle set as 5 sec.
	\item Dataset VI $-$ A cardiac perfusion dataset acquired during free-breathing. Available from the computational biomedical imaging group at the University of Iowa \footnote{\label{ft:uiowa}https://research.engineering.uiowa.edu/cbig/}. It was acquired using FLASH sequence on a 3T Siemens scanner (TR/TE=2.5/1.5ms) with a matrix size of $190 \times 90 \times 70$.
	\item Dataset VII $-$ A single-coil brain perfusion dataset acquired from multi slice 2-D dynamic susceptibility contrast (DSC)$^1$ with frame separation of TR=2s,  and with a matrix size of $128 \times 128 \times 60$.
\end{itemize}
All the measurements and reconstructions in this section were taken from these datasets. All tests and comparisons were run under the same conditions in a CPU-based implementation.


\subsection{Evaluation Scheme}
We validate our approach based on both qualitative and quantitative analyses. Whilst the qualitative analysis is derived from the observations and interpretations of physicians, for the quantitative analysis, we rely on computing a set of metrics between the gold-standard and the reconstructed MR image. From now, we refer to the reconstruction obtained from computing the sum-of-squares (SoS) using fully sampled data as {gold-standard}. More precisely, we base our findings on the following evaluation scheme.

\begin{figure*}[hbtp]
\centering
\includegraphics[width=1\textwidth]{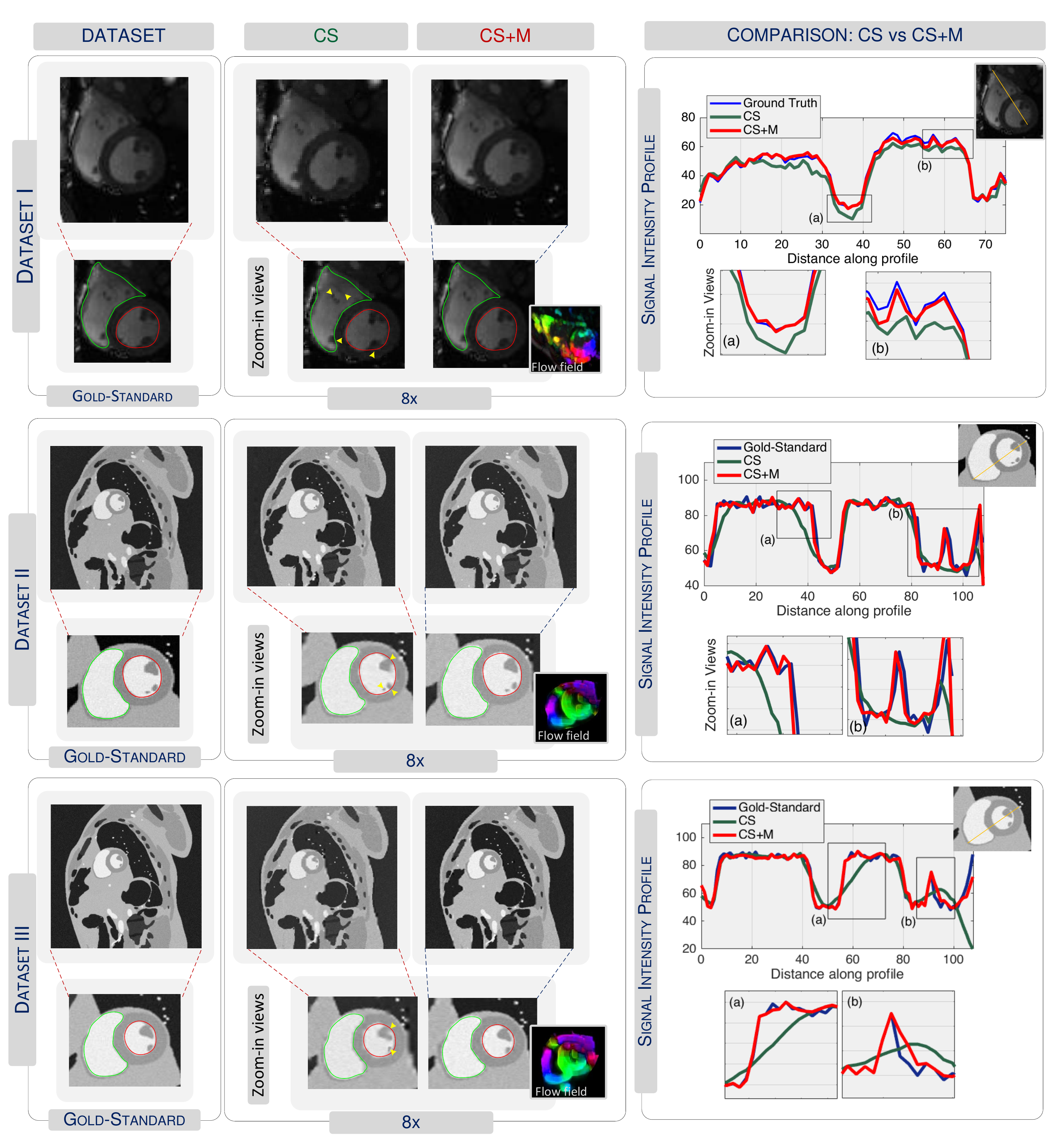}
\caption{Displayed reconstruction of three samples. (From left-to-right and top-to-bottom) Reconstructed fully-sampled data (i.e. \textcolor{black}{gold-standard}), reconstructed samples using CS and our CS+M. Zoom-in views in the middle part show more details in which the left (red) and right (green) ventricular endocardial borders have been outlined. Yellow arrows show artefacts created during the CS reconstruction. Signal intensity profiles retrieved from the reconstructed sample: blue lines in the plots show ideal signal intensity, green and red lines represent the CS and CS+M approaches. The results show that measurements generated from our approach are closer to the gold-standard.}
\label{fig::CSvsCSPlusM}
\end{figure*}

\begin{figure*}[!t]
\centering
\includegraphics[width=0.9\textwidth]{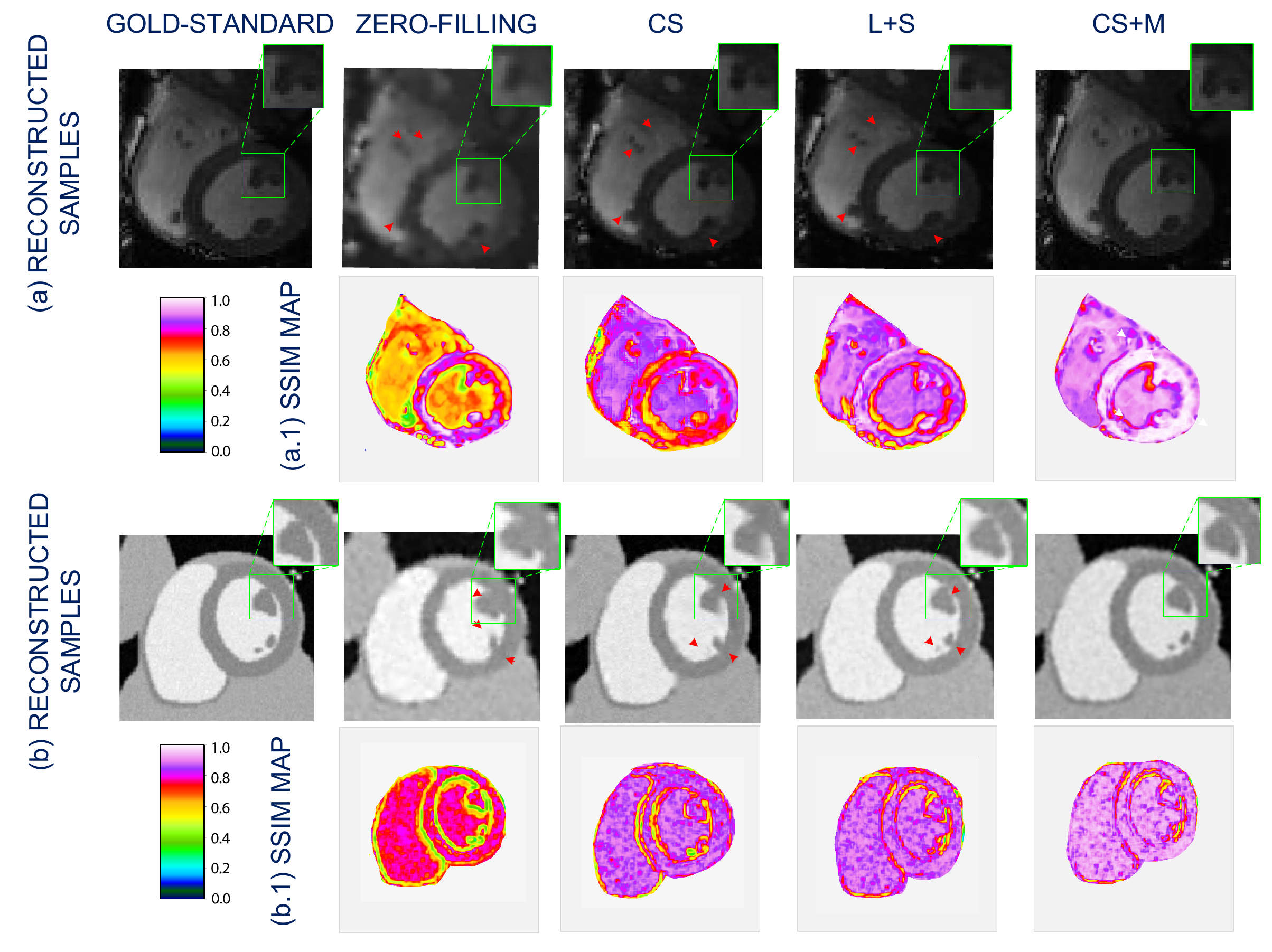}
\caption{Comparison performance of our approach and three reconstruction schemes along with the ground truth. (a) and (b) reconstructed samples from  datasets I and II in which loss of detail\textcolor{black}{s} and contrast, and introduction of blurring artefacts are pointed out with the red arrows. (a.1) and (b.1) \textcolor{black}{correspond to} SSIM maps, in which values as closer to 1 \textcolor{black}{means} better reconstruction quality. }
\label{fig::res4b}
\end{figure*}

\begin{figure*}[!t]
\centering
\includegraphics[width=1\textwidth]{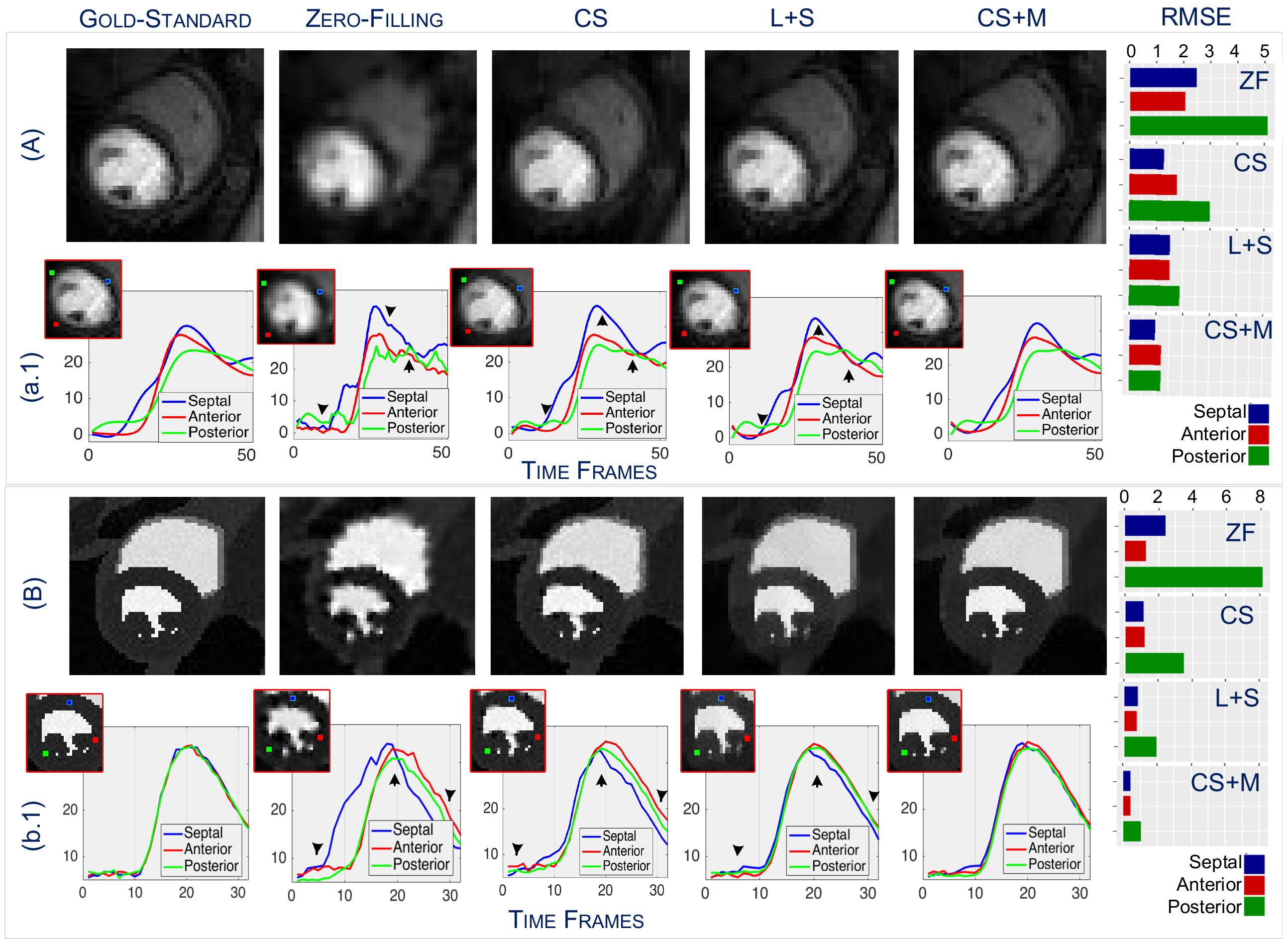}
\caption{Comparison performance on the temporal domain of our approach against three from the body of literature \textcolor{black}{(ZF, CS and L+S)}. (A) and (B) display a reconstructed sample for each evaluated scheme along with the gold-standard using Datasets V and VI. (a.1) and (b.1) show the corresponding temporal signal intensity profiles and error plots of three regions of interest at myocardium. \textcolor{black}{The RMSE metric in three different regions namely Septal, Anterior and Posterior, is displayed on the right hand side of the figure .}}
\label{fig::res5a}
\end{figure*}

\begin{figure*}[!t]
\centering
\includegraphics[width=0.95\textwidth]{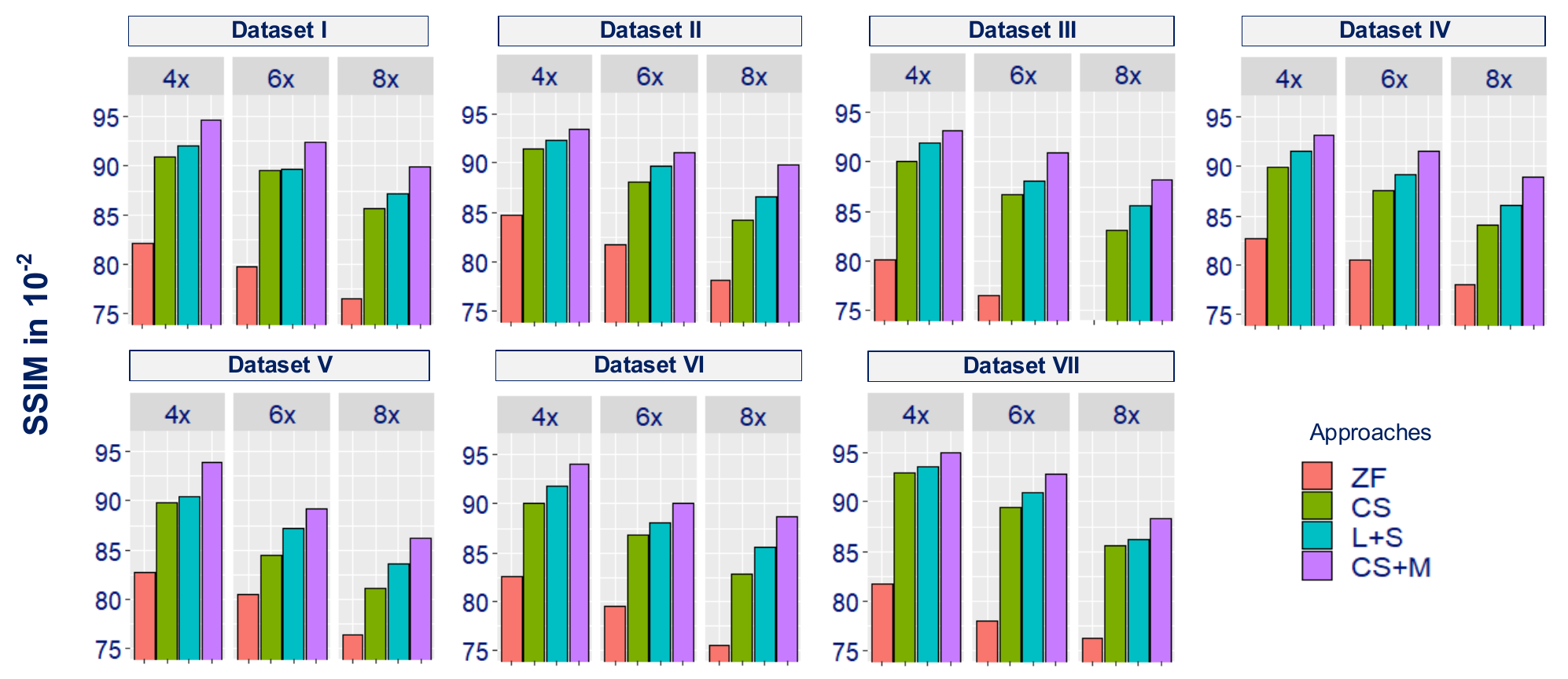}
\caption{Global performance analyses from our approach and three other reconstruction schemes \textcolor{black}{(ZF, CS and L+S)}. Displayed \textcolor{black}{SSIM} measures averaged over all the corresponding dataset denoted in $10^{-2}$ .}
\label{fig::res1d}
\end{figure*}

\begin{figure}[!t]
\centering
\includegraphics[width=0.35\textwidth]{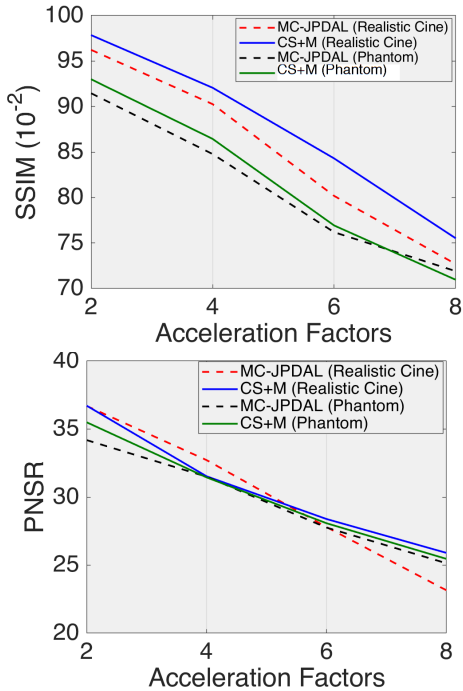}
\caption{Performance comparison of our approach vs another joint model MC-JPDAL \textcolor{black}{using the SSIM and PSNR metrics for two datasets}.}
\label{fig::plot}
\end{figure}

\begin{enumerate}[label=(\subscript{$E$}{\arabic*})]
\itemsep0em
\item Comparison of reconstructions between CS and CS+M for a range of acceleration factors associated to rates of undersampling in \textit{k}--space data. This comparison is based on visual comparisons of the 2D MR reconstructions at specific time steps and of selected 1D signal intensity profiles: Fig.~\ref{fig::CSvsCSPlusM}.
\item Careful analysis of the temporal reconstruction quality for different acceleration factors between CS, L+S and CS+M using the SSIM and SLMSE image-quality metrics: Fig.~\ref{fig::Drawing5}.
\item Visual assessment of the reconstruction quality,  and numerical visualisation of the intensity profiles and the LV area of our approach and three reference techniques for reconstruction \textcolor{black}{namely} zero-filling, CS~\citep{Lusting::2007} and L+S~\citep{Otazo::2015}: Fig.~\ref{fig::res4b}.
\item Temporal performance comparison of the reconstructions  between the CS+M and three reconstruction schemes (ZF, CS and L+S) from the literature along with the analysis of three regions of interest: Fig.~\ref{fig::res5a}.
\item Generalisation capabilities and global analysis performance of our approach vs. three reconstruction schemes \textcolor{black}{(ZF, CS and L+S)} and for different acceleration factors:  Fig.~\ref{fig::res1d}
\item Comparison of our approach vs MC-JPDAL~\citep{Zhao::2019}, which follows similar philosophy \textcolor{black}{as} our technique, that is- a joint model. Visual evaluation is reported in Fig.~\ref{fig::compJoint} whilst metric wise performance is displayed in Fig.~\ref{fig::plot}.
\end{enumerate}


We address the quantitative analysis relying on two well-established image-quality metrics: (i) the Structural Similarity (SSIM) Index~\citep{Wang::2004}, which calculates the similarity of two image reconstructions from the contrast, luminance and structure, 
and (ii) the inverted Localised Mean Squared Error (sLMSE)~\citep{Grosse::2009} that computes the local similarity based on local patches.  The computation of the sLMSE is a normalised and inverted measure such that closer to 1 means higher quality reconstruction.

\subsection{Parameter Selection}
Parameters of our approach and the ones we compare with were set individually chosen from a range of values with respect to the best SSIM and sLMSE, for each type of data. In particular for the case of the L+S approach, we define the range of values as suggested in~\citep{Otazo::2015}  along with the code provided \textcolor{black}{by} the authors.

For our approach, there are three parameters to set $\gamma, \delta \text{ and } \beta$. These parameters were tested in a range of values and for each clinical application type as follows. For the parameter  $\beta \in [0.1,1]$, we found the best outcome by setting it to $0.45$ whilst for the other two parameters, $\gamma \text{ and } \delta$ , we tested them in the  range $[10^{-2},...,0.9]$.
It is to be noticed that the parameters of the CS+M model can be easily fine-tuned for a particular clinical application and kept fixed for other datasets with similar dynamic information. Moreover, we used TV \citep{Rudin::1992} as an operator for promoting sparse representation in our and compared schemes. The objective of this work is to have a proof of concept to show the potentials and generalisation capabilities of our approach for MRI reconstruction. Therefore, a detailed investigation \textcolor{black}{on regularisers} shall be tackled in future work.


\subsection{Results and Discussion}
We describe our findings following the scheme described in Section 3.2.  We start by evaluating classic CS MRI scheme against our CS+M model. In Fig.~\ref{fig::CSvsCSPlusM}, we display three reconstructed samples with an acceleration factor of 8x, and for three different datasets. Visual assessment of the reconstructed samples agrees with our initial hypothesis that the incorporation of motion in the reconstruction model benefits the output quality.
Most notably, it can be observed in a comparison with the gold-standard,  \textcolor{black}{in} the right and left ventricular endocardial  borders (see outlined green and red regions), that CS+M offers better reconstructions in terms of contrast and shape than the CS reconstructions (see yellow arrows). Moreover, in a closer inspection at the zoomed-in views, it is to be noticed the loss of fine details and blurring effect at the papillary muscles.

This is further reflected in the signal intensity profile\textcolor{black}{s on} \textcolor{black}{the} right side of Fig.~\ref{fig::CSvsCSPlusM}, where, and for all the displayed cases, the CS+M approach (red line) is closer to the gold-standard (blue line). Although CS based reconstruction (green line) offers a good approximation to the gold-standard, it fails to eliminate all perturbations such as blurring artefacts (see yellow arrows), \textcolor{black}{contrary to our approach}. 
This is reflected in the behaviour of signal intensity profile in which significant \textcolor{black}{oscillations} are observed; as it is visible on the zoomed-in views at \textcolor{black}{the} right side of Fig.~\ref{fig::CSvsCSPlusM}.
In particular, it can be observed \textcolor{black}{in} the signal intensity profile of Dataset III,  acquired during free-breathing (last row of Fig.~\ref{fig::CSvsCSPlusM}), that signals generated from CS describe \textcolor{black}{strong oscillations compared to the gold-standard and CS+M ones} yielding to an unstable behaviour (see zoom-in views).

But $-$ Is there a significant difference in reconstruction quality between our approach and the compared reconstruction schemes?
To respond to this question we compute the reconstruction of  all datasets under high undersampling factors up to 12x. The plot of Fig.~\ref{fig::Drawing5} shows the SSIM and sLMSE curves for CS, L+S and our approaches where we can observe that the CS+M outperforms CS and L+S at all acceleration counts. For example, with an acceleration of 8x, the reconstruction quality obtained with our approach can only be achieved with an acceleration of 6x for CS for both metrics. Meaning that, despite reducing the measurement samples, our approach is still able to generate higher-quality images. A similar behaviour, but with a less numerical difference, is observed in a comparison between our approach and L+S. \textit{We then underline a strength of our approach that is its performance even with highly acceleration factors.}

For a more detailed analysis, we evaluate our approach by comparing its performance against three different reconstruction schemes: zero-filling, CS and L+S. Fig.~\ref{fig::res4b} displays reconstructed samples, taken from datasets I and II and undersampled at 8x, of the chosen schemes and our proposed one along with the gold-standard. By visual evaluation, we observe that the compared schemes tend to produce blurring artefacts and \textcolor{black}{to lose} fine details (see red arrows at (a), (b) of Fig.~\ref{fig::res4b} \textcolor{black}{ and the zoom-in views in the green squares)}. This is further reflected and better appreciated in the displayed SSIM maps at (a.1), (b.1) from the same figure, which offer meaningful comparison of local image quality over space. A closer inspection shows that our approach produces a reconstruction with higher similarity metric to the gold-standard.


Next, we investigate - how both CS+M and the compared schemes perform in the temporal domain.
We execute another experiment using the whole cardiac perfusion datasets\textcolor{black}{. We} select this application since the dynamic information contained in it differs from the cardiac cine information. In Fig.~\ref{fig::res5a}, we show four reconstructed samples resulted from the compared schemes and our approach along with the
gold-standard, (A) at 7.5x and (B) at 8x acceleration. By visual inspection, we observe that the effects discussed above from the cardiac cine datasets also prevail for the perfusion cardiac case, meaning that, the compared approaches reflect perturbations in the reconstructions such as blurring effects, and loss of contrast and details \textcolor{black}{especially visible in the right and left ventricular endocardium}.

\begin{figure*}[!t]
\centering
\includegraphics[width=1\textwidth]{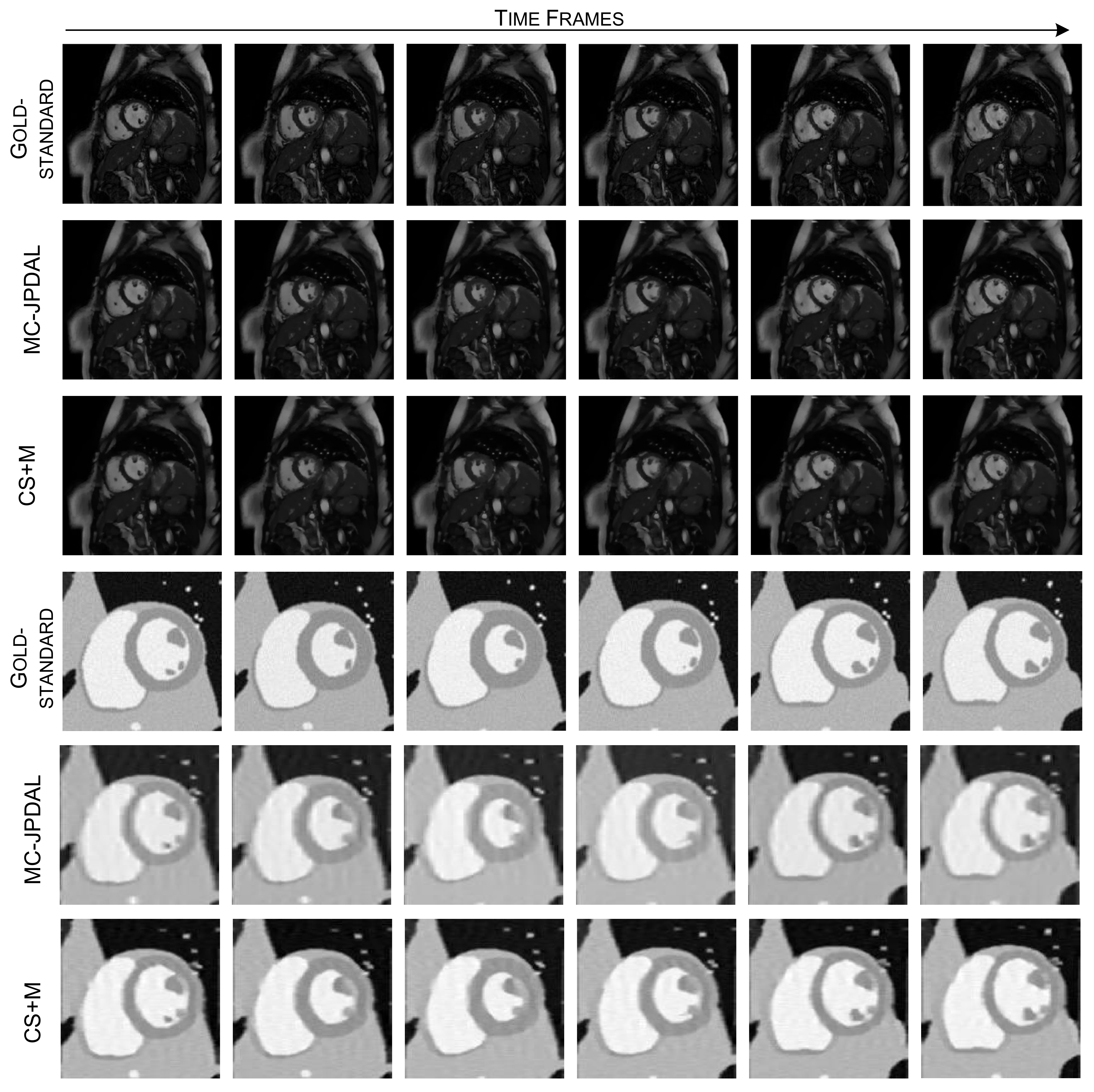}
\caption{\textcolor{black}{Visual comparison of our technique vs another joint model (MC-JPDAL) on reconstructed time samples.} The first dataset  was undersampled with a golden angle of 10 lines whilst the second  with a cartesian one with 6x. \textcolor{black}{Columns } display the same time reconstructed frame along with the gold-standard.}
\label{fig::compJoint}
\end{figure*}

We also computed the reconstructions over the whole corresponding datasets, and extracted the temporal intensity for three regions of interest in the myocardium. The results are displayed at Fig.~\ref{fig::res5a} - (a.1),(b.1). A closer inspection show\textcolor{black}{s} higher temporal fluctuations which translate in higher residual artefacts for the compared schemes \textcolor{black}{(see blue arrows)}, whilst our approach \textcolor{black}{gives} a more stable signal which is in fact closer to the gold-standard in all cases. This is further reflected by the error plots (Root-mean-square error RMSE \textcolor{black}{computed in the three identified regions of interest}), in which our approach has the lowest error value for all reconstructed samples. From the compared schemes, the one that performs better is L+S, however, we observe on the temporal intensity signals and the resulted reconstructions (Figs.~\ref{fig::res4b}, \ref{fig::res5a}), contrast variations, \textcolor{black}{blurring artefacts} and oscillations in the signal intensity. By contrast, our approach display\textcolor{black}{s} more stable signal intensities with less fluctuations and blurring artefacts, and better preservation of the shape. 



For quantitative evaluation of generalisation capabilities of our approach and for a global analysis performance, we report the results in Fig.~\ref{fig::res1d}. The reported numbers are the average of the selected image metrics across the entire corresponding datset. It is to be noticed, and as we previously mentioned, both metrics as closer to 1 as higher reconstruction quality. From the results, we observe that our approach outperforms the compared reconstruction schemes with respect to the SSIM metric. We also notice that, in the case where there are different types of dynamics - for example  the organ's physiological motion along with the breathing - the reconstruction benefits from this knowledge reflecting in a significant improvement of the reconstruction (for example see Datasets V and VI). Overall, our approach reports the highest values for all datasets and for all acceleration factors. This shows the ability of our approach to generalise to different types of dynamics, e.g. free-breathing and perfusion data.

We finally compare our approach against~\citep{Zhao::2019}, which is a recent technique with similar philosophy \textcolor{black}{as} ours.  We include an extra cine cardiac dataset provided in ~\citep{Zhao::2019} (see that paper for \textcolor{black}{a} detailed description of the dataset) composed of 30 temporal frames of $256\times 256$. We use
an undersampling golden angle radial pattern with 15 rays whilst the second dataset is undersampled with a cartesian patter of 6x. We report the results in Figs.~\ref{fig::compJoint}, \ref{fig::plot}.
We start by displaying a set of visual \textcolor{black}{time reconstruction} outputs in Fig.~\ref{fig::compJoint}.
In a closer inspection of the outputs, we can see that  MC-JPDAL  reconstructions exhibit blurry artefacts and tend to lose the initial contrast. These effects are significantly less noticeable with our method.  See for example,
the outputs at rows 4th-6th , where the \textcolor{black}{left ventricular} blood pool displays loss of details, this is also observed in the first rows outputs.
We further support the visual evaluation by reporting a metric wise comparison in Fig.~\ref{fig::plot}. The displayed  plots  are  the average of each metric for all the datasets and for different acceleration factor. With respect to the SSIM we outperform MC-JPDAL for the two compared datasets. Whilst for the  PNSR we readily compete with that approach but specially we perform better with high acceleration factors.

We remark that although the MC-JPDAL~\citep{Zhao::2019} is based on similar philosophy \textcolor{black}{as} ours, the improvement of our approach comes from the fact that the technique of that~\citep{Zhao::2019}
simplifie\textcolor{black}{s} the weighted optical flow task by considering
only affine displacement fields whereas, in our model,
the displacement field is not restricted to any particular class of functions.



Overall, a global inspection of the qualitative and quantitative analyses from the compared schemes shows the following drawbacks, on which our CS+M approach improves:

\begin{itemize}
\itemsep0em
  \item[$\checkmark$] \textit{Introduction of Temporal Artefacts.} We notice that the compared schemes exhibit significant  {oscillations} in the temporal signal intensity, which is translated in higher residual aliasing \textcolor{black}{and blurring} artefacts. This effect is prevailed for all datasets. By contrast, the CS+M approach displayed  temporal stable signals which are closer to the gold-standard yielding to higher reconstructions quality.
  \item[$\checkmark$] \textit{Fine Details and Shape Preservation.}  From the results, we observe that compared approaches exhibit loss in details and blurring effect, for example  the papillary muscles of the heart or in the brain, and even more significantly noticed during complex dynamic transitions such as the contraction-expansion of the heart. Conversely, our approach offered a better spatio-temporal fidelity for all applications.
  \item[$\checkmark$] \textit{Stability under Physiological Motion.}  We notice that the compared schemes tend to have significant fluctuations  during changes in the dynamic information such as systolic phases (in the case of the heart) or when additional dynamic information appears as in the case of free-breathing datasets. During these events, the results coming from our scheme exhibit a more stable behaviour resulting in a reconstruction closer to the gold-standard.
\end{itemize}

These points comes to highlight the benefit of incorporating motion into the algorithmic MRI reconstruction whilst supporting our initial hypothesis.

\section{Conclusion}
In this paper we address a central question in MRI which is $-$ How to get high quality MRI reconstructed images under highly undersampling factors? We respond to this question through a new approach called CS+M, in which the novelty largely relies on incorporating, explicitly and simultaneously in a single model, an estimate of the scene's motion to the algorithmic MRI reconstruction to provide higher quality images with less motion artefacts. We demonstrate the potentials of our approach based on exhaustive qualitative and quantitative analyses, in which we show that our approach outperforms traditional reconstruction schemes in terms of preservation of fine details and organs' shape and \textcolor{black}{reduction of} blurring artefacts.
Whilst the objective of this work is to open a new line of research for further clinical investigation,
since the CS+M model proves that motion has significant positive effects that translates to clinical potentials, future work might address how to improve the motion estimation.




\bibliographystyle{model2-names.bst}\biboptions{authoryear}
\bibliography{mainSub}


\end{document}